\documentclass{article}
\usepackage[nonatbib, preprint]{neurips_2025}
\usepackage[utf8]{inputenc} 
\usepackage[T1]{fontenc}    
\usepackage{hyperref}       
\usepackage{url}            
\usepackage{booktabs}       
\usepackage{amsfonts}       
\usepackage{amsmath}
\usepackage{nicefrac}       
\usepackage{microtype}      
\usepackage{xcolor}         
\usepackage{graphicx}
\usepackage{comment}
\usepackage{multirow}
\usepackage{threeparttable}
\usepackage{tabularx}
\usepackage{arydshln}
\usepackage{placeins}
\usepackage{makecell}
\usepackage{threeparttablex}
\usepackage{arydshln}
\usepackage{svg}
\usepackage{adjustbox}
\usepackage{subcaption}
\usepackage{pythonhighlight}
\usepackage[acronym,shortcuts]{glossaries}
\usepackage[maxcitenames=1, url=false, sortcites]{biblatex}
\usepackage{listings}
\lstdefinestyle{pycharm}{
  language=Python,
  basicstyle=\ttfamily\small,
  keywordstyle=[1]\color{blue}\bfseries,
  keywordstyle=[2]\color{cyan}\bfseries,
  keywordstyle=[3]\color{violet}\bfseries,
  stringstyle=\color{green!70!black},
  commentstyle=\color{gray},
  morecomment=[l][\color{orange}]{@}
  showstringspaces=false,
  numbers=right,
  numberstyle=\tiny\ttfamily\color{gray},
  numbersep=5pt,
  tabsize=2,
  breaklines=true,
  morekeywords=[1]{import,from,class,def,return,if,elif,else,for,in,while,with,as,is,not,or,and,pass,break,continue,global,nonlocal,try,except,finally,raise,assert},
  morekeywords=[2]{True,False,None},
  morekeywords=[3]{int,float,str,list,tuple,dict,set,bool},
  emph={self},
  emphstyle=\color{red}\bfseries,
  basicstyle=\ttfamily
}

\usepackage{array}
\usepackage[tableposition=top]{caption}
\usepackage{colortbl}
\usepackage{rotating}
\usepackage{tabularray}
\usepackage{orcidlink}
\DeclareCaptionFormat{custom}
{
\small\textsc{#1#2}#3
}
\usepackage{pifont}
\newcommand{\cmark}{\ding{51}}%
\definecolor{lightergray}{gray}{0.95}
\newcommand{\xmark}{{\color{lightergray}\ding{55}}}%
\svgpath{{../figures/}}

\definecolor{Gray}{gray}{0.9}
\newcolumntype{g}{>{\columncolor{Gray}}c}

\def\BibTeX{{\rm B\kern-.05em{\sc i\kern-.025em b}\kern-.08em
    T\kern-.1667em\lower.7ex\hbox{E}\kern-.125emX}}
\usepackage{pifont}
\definecolor{lightergray}{gray}{0.95}

\begin{document}

\newacronym{icu}{ICU}{Intensive Care Unit}
\newacronym{ehr}{EHR}{Electronic Health Record}

\newacronym{ml}{ML}{Machine Learning}
\newacronym{dl}{DL}{Deep Learning}


\newacronym{cnn}{CNN}{Convolutional Neural Network}
\newacronym{fnn}{FNN}{Feed-forward Neural Network}
\newacronym{rnn}{RNN}{Recurrent Neural Network}
\newacronym{lstm}{LSTM}{Long Short-Term Memory}
\newacronym{gru}{GRU}{Gated Recurrent Unit}
\newacronym{tcn}{TCN}{Temporal Convolutional Network}

\newacronym{da}{DA}{Domain Adaptation}
\newacronym{sda}{SDA}{Supervised Domain Adaptation}
\newacronym{uda}{UDA}{Unsupervised Domain Adaptation}
\newacronym{dg}{DG}{Domain Generalization}

\newacronym{yaib}{YAIB}{Yet Another ICU Benchmark}

\newacronym{aki}{AKI}{Acute Kidney Injury}
\newacronym{los}{LoS}{Length of Stay}
\newacronym{format}{SICUD}{Standardized ICU Definition}

\newacronym{mar}{MAR}{missing at random}
\newacronym{mcar}{MCAR}{missing completely at random}
\newacronym{mnar}{MNAR}{missing not at random}

\newacronym{locf}{LOCF}{last observation carried forward}

\newacronym{auroc}{AUROC}{Area Under the Reciever Operating Characteristic}
\newacronym{auprc}{AUPRC}{Area Under the Precision Recall Curve}
\newacronym{mimic}{MIMIC}{Medical Information Mart for Intensive Care}
\newacronym{miiv}{MIMIC-IV}{Medical Information Mart for Intensive Care IV}
\newacronym{miiii}{MIMIC-III}{Medical Information Mart for Intensive Care III}
\newacronym{eicu}{eICU}{eICU Collaborative Research Database}
\newacronym{hirid}{HiRID}{High Time Resolution ICU Dataset}
\newacronym{aumc}{AUMCdb}{AmsterdamUMCdb}
\newacronym{sssd}{SSSD}{Structured State-Space Diffusion}
\newacronym{csdi}{CSDI}{Conditional Score-based Diffusion Model for Probabilistic Time Series Imputation}
\newacronym{ae}{AE}{Autoencoder}
\newacronym{amc}{AMC}{AmsterdamUMCdb}
\newacronym{assm}{ASSM}{Attention-Based Sequence-to-Sequence Model}
\newacronym{bi-gan}{Bi-GAN}{Bi-Directional Generative Adversarial Network}
\newacronym{bidmc}{BIDMC}{Beth Israel Deaconess Medical Center}
\newacronym{bo}{BO}{Blackout}
\newacronym{brnn}{BRNN}{Bidirectional Recurrent Neural Network}
\newacronym{brits}{BRITS}{Bidirectional Recurrent Imputation for Time Series}
\newacronym{crps}{CRPS}{Continuous Ranked Probability Score}
\newacronym{csdi_t}{CSDI\_T}{Conditional Score-based Diffusion Models for Tabular data}
\newacronym{dacmi}{DACMI}{Data Analytics Challenge on Missing data Imputation}
\newacronym{dbp}{dbp}{diastolic blood pressure}
\newacronym{dm}{DM}{Diffusion Model}
\newacronym{dmsa}{DMSA}{Diagonally-Masked Self-Attention}
\newacronym{e2gan}{E2GAN}{End-to-End Generative Adversarial Network}
\newacronym{elbo}{ELBO}{evidence lower bound}
\newacronym{em}{EM}{Expectation Maximization}
\newacronym{gain}{GAIN}{Generative Adversarial Imputation Networks}
\newacronym{gan}{GAN}{Generative Adversarial Networks}
\newacronym{gp}{GP}{Gaussian Processes}
\newacronym{hippa}{HIPPA}{Health Insurance Portability and Accountability Act}
\newacronym{hr}{hr}{heart rate} 
\newacronym{jsd}{JSD}{Jensen-Shannon Divergence}
\newacronym{kld}{KLD}{Kullback-Leibler Divergence}
\newacronym{knn}{KNN}{k-nearest neighbor}
\newacronym{lgbm}{LGBM}{Light Gradient-Boosting Machine}
\newacronym{lime}{LIME}{Local Interpretable Model-agnostic Explanations}
\newacronym{lls}{LLS}{Local Least Square}
\newacronym{map}{map}{mean arterial pressure}
\newacronym{mae}{MAE}{Mean Absolute Error}
\newacronym{mcmc}{MCMC}{Monte Carlo Markov Chain}
\newacronym{mf}{MF}{MissForest}
\newacronym{mi}{MI}{Multiple Imputation}
\newacronym{mice}{MICE}{Multivariate Imputation by Chained Equations}
\newacronym{mit}{MIT}{Masked Imputation Task}
\newacronym{mlp}{MLP}{multi-layer perceptron}
\newacronym{mre}{MRE}{Mean Relative Error}
\newacronym{mrnn}{MRNN}{Multi-Directional Recurrent Neural Network}
\newacronym{mse}{MSE}{Mean Square Error}
\newacronym{naa}{NAA} {Neighborhood-Aware Autoencoder}
\newacronym{nf}{NF}{Normalizing Flow}
\newacronym{np}{NP}{Neural Process}
\newacronym{o2sat}{o2sat}{oxygen saturation}
\newacronym{ort}{ORT}{Observed Reconstruction Task}
\newacronym{relu}{ReLU}{rectified linear unit}
\newacronym{resp}{resp}{respiratory rate}
\newacronym{rmse}{RMSE}{Root Mean Square Error}
\newacronym{saits}{SAITS}{Self-Attention-based Imputation for Time Series}
\newacronym{sbp}{sbp}{systolic blood pressure}
\newacronym{shap}{SHAP}{SHapley Additive exPlanations}
\newacronym{si}{SI}{Single Imputation}
\newacronym{sota}{SOTA}{state-of-the-art}
\newacronym{ssgan}{SSGAN}{Semi-Supervised Generative Adversarial Network}
\newacronym{sting}{STING}{Self-attention based Time-series Imputation Networks using GAN}
\newacronym{vae}{VAE}{Variational Autoencoder}
\newacronym{wgan}{WGAN}{Wasserstein Generative Adversarial Network }
\newacronym{xai}{XAI}{Explainable Artificial Intelligence}
\newacronym{cinc}{CINC}{Computing in Cardiology Challenge}
\glsdisablehyper
\newcommand\mycite[1]{\AtNextCite{\defcounter{maxnames}{1}}\citeauthor{#1}}
\newacronym{imp}{\textsl{ICI}}{\textsl{Intensive Care Imputer}}
\title{
Closing Gaps:\\
An Imputation Analysis of ICU Vital Signs
}
\author{
  Alisher Turubayev*$^1$\orcidlink{0000-0001-7721-7409},
  Anna Shopova*$^1$\orcidlink{0009-0008-0627-9571}, 
  Fabian Lange*$^1$, \\
  \textbf{
  Mahmut Kamalak*$^1$, 
  Paul Mattes*$^1$,
  Victoria Ayvasky*$^1$, 
  Bert Arnrich$^1$,}\\
\textbf{Bjarne Pfitzner$^1$,
  Robin P. van de Water$^{1\dagger}$\orcidlink{0000-0002-2895-4872}}\\
  * Authors contributed equally. $\dagger$ Supervised this work. \\
 $^1$ Hasso Plattner Institute, University of Potsdam, Germany\\
\texttt{Corresponding author email: robin.vandewater@hpi.de}}

\maketitle

\begin{abstract}
As more Intensive Care Unit (ICU) data becomes available, the interest in developing clinical prediction models to improve healthcare protocols increases. However, lacking data quality still hinders clinical prediction using Machine Learning (ML). Many vital sign measurements, such as heart rate, contain sizeable missing segments, leaving gaps in the data that could negatively impact prediction performance. Previous works have introduced numerous time-series imputation techniques. Nevertheless, more comprehensive work is needed to compare a representative set of methods for imputing ICU vital signs and determine the best practice. In reality, ad-hoc imputation techniques that could decrease prediction accuracy, like zero imputation, are still used. In this work, we compare established imputation techniques to guide researchers in improving clinical prediction model performance by choosing the most accurate imputation technique. We introduce an extensible, reusable benchmark with, currently, 15 imputation and 4 amputation methods created for benchmarking on major ICU datasets. We hope to provide a comparative basis and facilitate further ML development to bring more models into clinical practice. 

\textbf{Software Repository:} \url{https://github.com/rvandewater/YAIB}

\end{abstract}


\section{Introduction}
Real-world environments, such as medical treatment centers, often collect large amounts of data through administrative processes, tests, and monitoring equipment.
Although patients are monitored throughout their ICU stay, many practical issues can lead to data loss and missing data; data quality remains an ongoing challenge for clinical prediction modeling. 
Some of this is by design (an infrequent sample rate due to lack of medical necessity) or unintentional (monitoring devices lose connection).
Clinical prediction is a field that emerged from \acrfull{ehr} data, which was formerly used for administrative purposes. 
For example, models could predict mortality~\cite{bakerContinuousAutomaticMortality2020} or sepsis~\cite{moorEarlyRecognitionSepsis2019} of patients within the ICU. 
To utilize \acrshort{ehr}s to create useful clinical models, the standard practice in \acrshort{ml} is to use imputation methods, which are algorithms to fill in missing data.
Similarly to the increase of advanced methods in \acrshort{ml} and \acrfull{dl} in recent years, the field of imputation has also rapidly developed, leaving practitioners and researchers with the question of which technique to use~\cite{zainuddinTimeSeriesData2022}.
Another important consideration is that there could be a difference in the performance of imputation methods when considering downstream tasks like the aforementioned sepsis and mortality predictions~\cite{shadbahrClassificationDatasetsImputed2022,zhangFalsePositiveFindings2021}.

Our work aims to provide researchers and healthcare practitioners with a deeper understanding of the performance of these \acrshort{ml} and \acrshort{dl} imputation models. 
We achieve this by utilizing three large, open-access ICU datasets with four types and three quantities of missingness in vital signs as our data source.
In total, we benchmark 15 different models for imputation.
This selection includes recently introduced generative diffusion-based models and attention-based techniques alongside more traditional statistical imputation methods.
The methods are incorporated within a benchmarking framework to allow for the replication of our experiments and reuse on current and future datasets. 
This setup allows researchers to build upon our results and create, benchmark, and compare imputation methods on standard ICU datasets.
Lastly, we provide a straightforward experiment pipeline to apply imputation to any clinical dataset of choice. 
\section{Background \& Related Work}
\subsection{Types of missingness}
We define \textit{missingness} as the absence of data where it is unexpected. 
For example, an absence of data is expected when we consider time-series observations below the technical sampling frequency of a vital sign recording device.
We identify four significant types of missingness in this work: \textit{\acrfull{mcar}}, \textit{\acrfull{mar}}, \textit{\acrfull{mnar}}, and \textit{\acrfull{bo}}. 
Each presented missingness type is implemented in our framework as an \textit{amputation} mechanism, as it removes data to assess imputation performance.

\textit{\acrshort{mcar}} refers to where missingness introduces no statistical bias. 
For example, in a clinical trial, a recording of a group of patients might be missing due to failing equipment.
\textit{\acrshort{mar}} does introduce a bias, but this bias is systematically related exclusively to observed data.
For example, we record the sex of the participant as a response to a survey, and men are less likely to respond. With \textit{\acrshort{mnar}}, data are missing, that are systematically related to unobserved factors (i.e., events not measured in the experiment). We again take the survey example but assume that sex is not recorded in this case; this gives us a scenario where it is hard to account for bias introduced by the missing participants' responses.
For \textit{\acrshort{bo}} an entire subset of data is missing for several features during several timesteps. This type of missingness is particularly present in time-series and multimodal datasets, where entire modalities can be missing. This type of missingness is less commonly tested for but has a basis in existing literature~\cite{alcarazDiffusionbasedTimeSeries2023}.

\subsection{Datasets}
We used three major ICU datasets in our work, provided in the \acrfull{yaib}~\cite{vandewaterAnotherICUBenchmark2023} experiment framework. \acrshort{yaib} already implements a flexible framework for downstream tasks using most open-access ICU datasets and was chosen as an extensible foundation for the current work.
The MIMIC-III dataset~\cite{johnsonMIMICIIIFreelyAccessible2016} is the most commonly used dataset in \acrshort{ml}-based prediction tasks~\cite{syedApplicationMachineLearning2021}. 
The newer MIMIC-IV (MIMIC-IV) includes several improvements, including more and newer patient records, as well as a revised structure that includes regular hospital information; therefore, we use this version for our experiments. 
The eICU Collaborative Research Database (eICU)~\cite{pollardEICUCollaborativeResearch2018} is the first sizable multi-center dataset, and contains information from ICU monitoring systems from 208 participating hospitals in the US.
The High tIme Resolution ICU dataset (HiRID) was collected at Bern University Hospital, Switzerland, and incorporates more observations than the other datasets~\cite{hylandEarlyPredictionCirculatory2020}, allowing for a more exhaustive analysis of predictive systems.
A more comprehensive overview can be found in \autoref{tab:dataset_comparison_extended} and \textcite{sauerSystematicReviewComparison2022}.
\paragraph {Choice of Features}
For our comparison of imputation methods, we chose six temporal vital signs that showed significantly lower missingness than other recorded variables for each dataset:\textit{ \ac{hr}, \ac{resp}, \ac{o2sat}, \ac{map}, \ac{sbp}, and \ac{dbp}}. The nature of ICU data recording likely causes this pattern: monitoring equipment is usually continuously attached and allows for non-invasive recording, whereas, for example, lab values are typically taken only a few times per day and involve manual labor. We select these variables from the 52 (4 static, 48 temporal) variables in the harmonized datasets (downsampled to one hour~\cite{vandewaterAnotherICUBenchmark2023}) to ensure that we have sufficient ground truth data to accurately assess the performance of any imputation method. \autoref{fig:correlation} shows the missingness correlation between these features; \autoref{fig:informative-missingness} explores informative missingness by comparing the population that died within the ICU with those who survived.

\begin{table*}[]
\small
    \centering
\begin{threeparttable}
    \caption{Overview of the implemented imputation methods.}
\begin{tabularx}{\textwidth}{>{\hsize=.02\hsize}c>{\hsize=.02\hsize}c>{\hsize=.10\hsize}l>{\centering\arraybackslash\hsize=.75\hsize}X>{\hsize=.10\hsize}X>{\hsize=.30\hsize\arraybackslash}X>{\centering\arraybackslash\leavevmode}X}
\toprule
& & \textbf{Abbreviation} & \textbf{Original Publication} & \textbf{Year} & \textbf{Source} & \textbf{Novelty} \\
\hline
\parbox[t]{2mm}{\multirow{4}{*}{\rotatebox[origin=c]{90}{\textbf{Baseline}}}}
& & Zero & - &  &  $\omega$ \cite{pedregosaScikitlearnMachineLearning2011}$^1$ & Baseline \\
& & Median & - &  & $\omega$ \cite{pedregosaScikitlearnMachineLearning2011}$^1$& Baseline \\
& & Mean & - &  & $\omega$ \cite{pedregosaScikitlearnMachineLearning2011}$^1$ & Baseline \\
& & MostFrequent & - &  & $\omega$ \cite{pedregosaScikitlearnMachineLearning2011}$^1$ & Baseline \\
\hline
\parbox[t]{2mm}{\multirow{2}{*}{\rotatebox[origin=c]{90}{\textbf{Algo.}}}}
& & MICE & \textcite{buurenMiceMultivariateImputation2011} & \citeyear{buurenMiceMultivariateImputation2011} & $\omega$ \cite{jarrettHyperImputeGeneralizedIterative2022}$^2$ &   Equation based benchmark\\
& & MissForest & \textcite{stekhovenMissForestNonparametricMissing2012} & \citeyear{stekhovenMissForestNonparametricMissing2012} & $\omega$ \cite{jarrettHyperImputeGeneralizedIterative2022}$^2$ & Random forest based\\ 
\hline
\parbox[t]{2mm}{\multirow{10}{*}{\rotatebox[origin=c]{90}{\textbf{Deep learning}}}}
& & MLP & \textcite{junninenMethodsImputationMissing2004} & \citeyear{junninenMethodsImputationMissing2004}& $\sigma$ & DL baseline\\
& & Neural Processes & \textcite{garneloNeuralProcesses2018} & \citeyear{garneloNeuralProcesses2018} & $\alpha$ \cite{Dupont_Neural_process_pytorch_2019}$^3$ & First Neural Processes \\
\cline{3-7}
& \parbox[t]{2mm}{\multirow{3}{*}{\rotatebox[origin=c]{90}{\textbf{RNN}}}}
 & BRITS & \textcite{caoBRITSBidirectionalRecurrent2018} & \citeyear{caoBRITSBidirectionalRecurrent2018} & $\omega$ \cite{du2023PyPOTS}$^5$ & Bidirectional RNN\\
& & GRU-D & \textcite{cheRecurrentNeuralNetworks2018} & \citeyear{cheRecurrentNeuralNetworks2018} & $\alpha$ \cite{ciniFillingApMultivariate2022}$^4$ & Bidirectional LSTM/GRU \\
& & M4IP & \textcite{shiDeepDynamicImputation2021} & \citeyear{shiDeepDynamicImputation2021} & $\alpha$ \cite{ciniFillingApMultivariate2022}$^4$ & State decay RGRU-D\\
\cline{3-7}
& \parbox[t]{2mm}{\multirow{2}{*}{\rotatebox[origin=c]{90}{\textbf{At.}}}}
 & Attention & \textcite{vaswaniAttentionAllYou2017} & \citeyear{vaswaniAttentionAllYou2017} & $\omega$ \cite{du2023PyPOTS}$^5$ & Attention-based approach \\
& & SAITS & \textcite{duSAITSSelfAttentionbasedImputation2022} & \citeyear{duSAITSSelfAttentionbasedImputation2022} & $\omega$ \cite{du2023PyPOTS}$^5$ & SOTA Attention-based \\
\cline{3-7}
& \parbox[t]{2mm}{\multirow{2}{*}{\rotatebox[origin=c]{90}{\textbf{Gen.}}}}
 & Diffusion & \textcite{hoDenoisingDiffusionProbabilistic2020} & \citeyear{hoDenoisingDiffusionProbabilistic2020} & $\sigma$ \cite{nicholImprovedDenoisingDiffusion2021}$^6$ & Prob. Diff. model\\
 & & CSDI & \textcite{tashiroCSDIConditionalScorebased2021} & 2022 & $\alpha$ \cite{tashiroCSDIConditionalScorebased2021}$^7$ & Conditional Diff. model\\

%
\bottomrule
\end{tabularx}
\tiny{$\omega$ Wrapper framework $\alpha$ Adapted using open-access code $\sigma$ Self-implemented based on paper description
    $^1$ \url{https://github.com/scikit-learn/scikit-learn} $^2$ \url{https://github.com/vanderschaarlab/hyperimpute} $^3$ \url{https://github.com/EmilienDupont/neural-processes} $^4$ \url{https://github.com/Graph-Machine-Learning-Group/grin} $^5$\url{https://github.com/WenjieDu/PyPOTS} $^6$ \url{https://github.com/diff-usion/Awesome-Diffusion-Models} $^7$ \url{https://github.com/ermongroup/CSDI}} 
\label{tab:imputation_methods}
\end{threeparttable}
\vspace{-0.5cm}
\end{table*}


\subsection{Imputation methods}
We conducted a systematic literature review to discover promising imputation technologies. 
\autoref{tab:imputation_methods} shows the imputation methods we benchmark in this work. We distinguish several categories, as shown on the left side of the table.
\textbf{Baseline} methods are still commonly used in many applications of data preparation for \acrshort{ml} modeling as they are deterministic and computationally cheap.  \textbf{Algorithmic} methods~\cite{buurenMiceMultivariateImputation2011, stekhovenMissForestNonparametricMissing2012} use statistical assumptions on the data to create iterative algorithms; the methods are robust for simpler data. We also include several \textbf{\acrshort{dl}} methods in our comparison, with the simplest being a Multilayer Perceptron-based imputation. \textbf{RNN}-based methods~\cite{caoBRITSBidirectionalRecurrent2018, cheRecurrentNeuralNetworks2018, shiDeepDynamicImputation2021} have traditionally performed well on time-series prediction and imputation. \textbf{Attention} methods~\cite{vaswaniAttentionAllYou2017, garneloNeuralProcesses2018, duSAITSSelfAttentionbasedImputation2022} have developed rapidly in the past years and shown potential for various prediction tasks. \textbf{Generative} imputation methods include 
the more recent diffusion models~\cite{hoDenoisingDiffusionProbabilistic2020, tashiroCSDIConditionalScorebased2021}. 
\paragraph{Medical time series imputation benchmarks}
We recognize several earlier attempts at collecting and benchmarking imputation methods \cite{jagerBenchmarkDataImputation2021, sunReviewDeepLearning2020, perez-lebelBenchmarkingMissingvaluesApproaches2022,shadbahrClassificationDatasetsImputed2022, psychogyiosMissingValueImputation2023, luoEvaluatingStateArt2022, jarrettHyperImputeGeneralizedIterative2022} (\autoref{tab:related-works-overview}). 
\textcite{jagerBenchmarkDataImputation2021} investigated the performance of six imputation methods; however, there was no medical time-series among the tested datasets.
\textcite{perez-lebelBenchmarkingMissingvaluesApproaches2022a} focused on "classical" \acrshort{ml} and algorithmic methods of imputation; they do not include any \acrshort{dl} imputation methods in the analysis. 
\textcite{psychogyiosMissingValueImputation2023} conducted a benchmark of imputation methods on a closed-source tabular dataset.
\textcite{luoEvaluatingStateArt2022} reports on a challenge; the investigated dataset and methods limit the applicability to current data. 
\textcite{jarrettHyperImputeGeneralizedIterative2022} introduce the HyperImpute framework; we provide a more comprehensive medical vital sign analysis~\cite{duaUCIMachineLearning2017}. 
\textcite{sunReviewDeepLearning2020} investigate nine imputation methods on medical datasets. 
However, they do not provide an open-access extensible framework for implementing the imputation methods in a downstream task. 
Finally, \textcite{shadbahrClassificationDatasetsImputed2022} benchmark five imputation methods; the choice of methods and datasets is limited compared to our approach.


Our work uses a variety of missingness patterns. It enables the evaluation of both the imputation and downstream classification tasks, ensuring that the benchmarks reflect real-world performance. Additionally, we utilize three \acrshort{icu} datasets and provide a transparent and publicly available experimental setup, ensuring reproducibility and enabling easier comparison of results. Future work includes more imputation methods (25 in total, see \autoref{tab:related-works-overview}), results once we have verified their implementation.




\section{Comparing imputation technique performance}
We provide a standardized interface for imputation methods. We have used this to \textit{1)} wrap interfaces of earlier frameworks~\cite{pedregosaScikitlearnMachineLearning2011, jarrettHyperImputeGeneralizedIterative2022}, \textit{2)} adapt open-source code to our interface~\cite{ciniFillingApMultivariate2022, alcarazDiffusionbasedTimeSeries2023, tashiroCSDIConditionalScorebased2021}, and \textit{3)} easily implement imputation methods without an existing code-base. 
We use the Hyperimpute~\cite{jarrettHyperImputeGeneralizedIterative2022} and PyPOTS~\cite{du2023PyPOTS} frameworks which implement a number of imputation methods. 
The rest of the methods are implemented directly in \acrshort{yaib}~\cite{vandewaterAnotherICUBenchmark2023}. 
Hyperparameter tuning was performed for the \acrshort{dl} and \acrshort{ml} methods (see \autoref{sec:hyperparameter_optimization_tables}). 
Both the methods from the imputation packages and the methods implemented within \acrshort{yaib} are wrapped in an \texttt{ImputationWrapper} interface, which derives from a Pytorch-lightning \texttt{DLWrapper} module (see \autoref{app:extensibility} to implement new imputation methods).
Additionally, we developed the \texttt{ampute\_data(missing\_type, missing\_amount)} functionality which generates the dataset with artificially introduced missing values and a boolean mask indicating their location. The code is provided in the appendix. Using this function, one can quickly generate datasets with different types and levels of missingness to test and evaluate current and future imputation methods.
We shortly describe the implementation of each missingness technique below.

The \textit{\acrshort{mcar}} amputation mechanism generates missing values randomly without considering any additional input from the data or its characteristics.
For the \textit{\acrshort{mar}} amputation mechanism, we select a subset of fully observed variables. Then, missing values are introduced to the remaining variables by a logistic model~\cite{mayerRmisstasticUnifiedPlatform2021}. 
The proportion of missing values in these variables is re-scaled to match the desired proportion of overall missingness. 
The \textit{\acrshort{mnar}} also utilizes a logistic masking model. 
We split the variables into a set of inputs for a logistic model and a set whose missing probabilities are determined by the logistic model. The coefficients for the logistic masking model are selected such that $W^\top x$ has a unit variance, where $W$ is the subset of observed variables, and $x$ is the corresponding missing variable~\cite{sportisseSpEcialitDoctorale}. 
Then, inputs are masked; the missing values from the second set will depend on the masked values. 
Finally, the \textit{\acrshort{bo}} implementation takes a proportion of missing values; the function randomly selects rows in the data matrix and sets all their values to missing.

We have chosen three metrics to demonstrate the individual aspects of each method: (1) The averaged error over every time-series \textit{\acrfull{mae}}, (2) penalize higher individual errors more strongly\textit{\acrfull{rmse}}, and (3) demonstrate if an imputation method is capable of reconstructing the original distribution \textit{\acrfull{jsd}} (lower is better for each).

\begin{figure}[ht]
    \centering
    \centering
    \includegraphics[width=\textwidth]{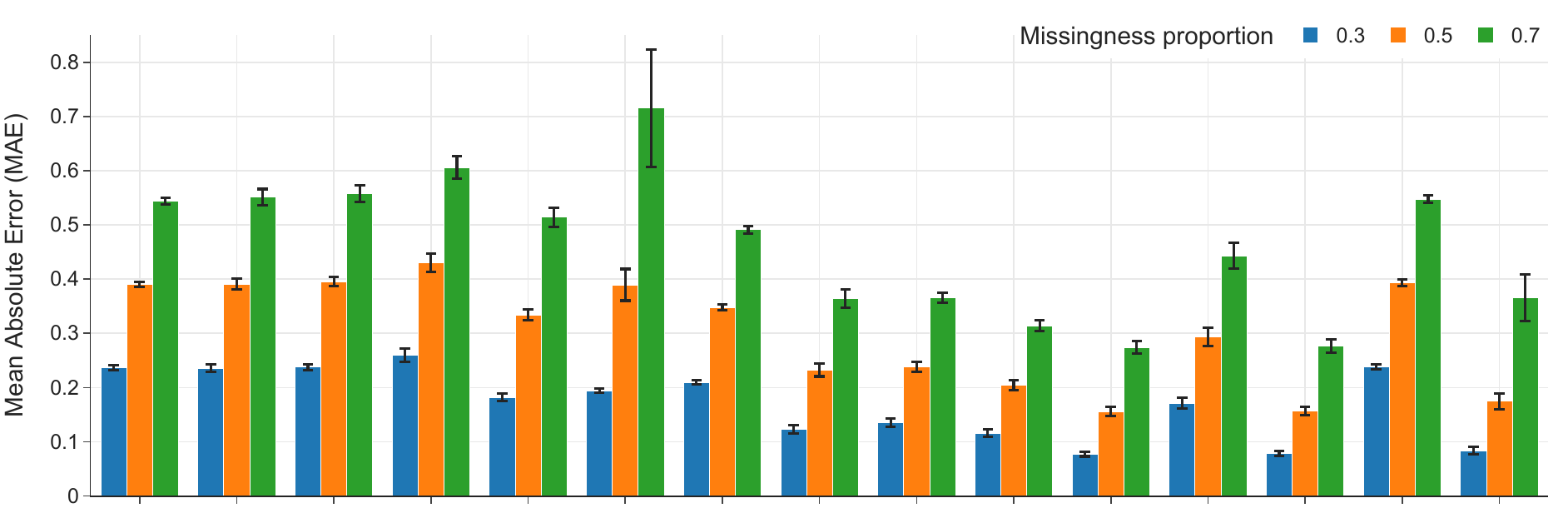}
    \includegraphics[width=\textwidth]{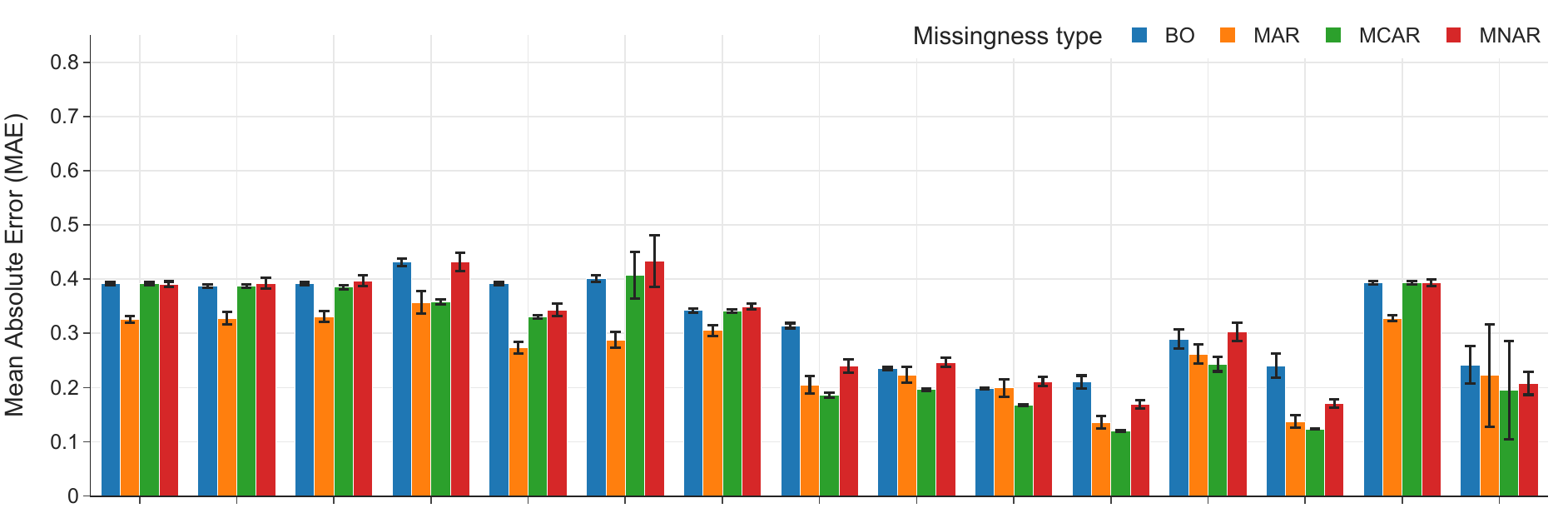}
    \includegraphics[width=\textwidth]{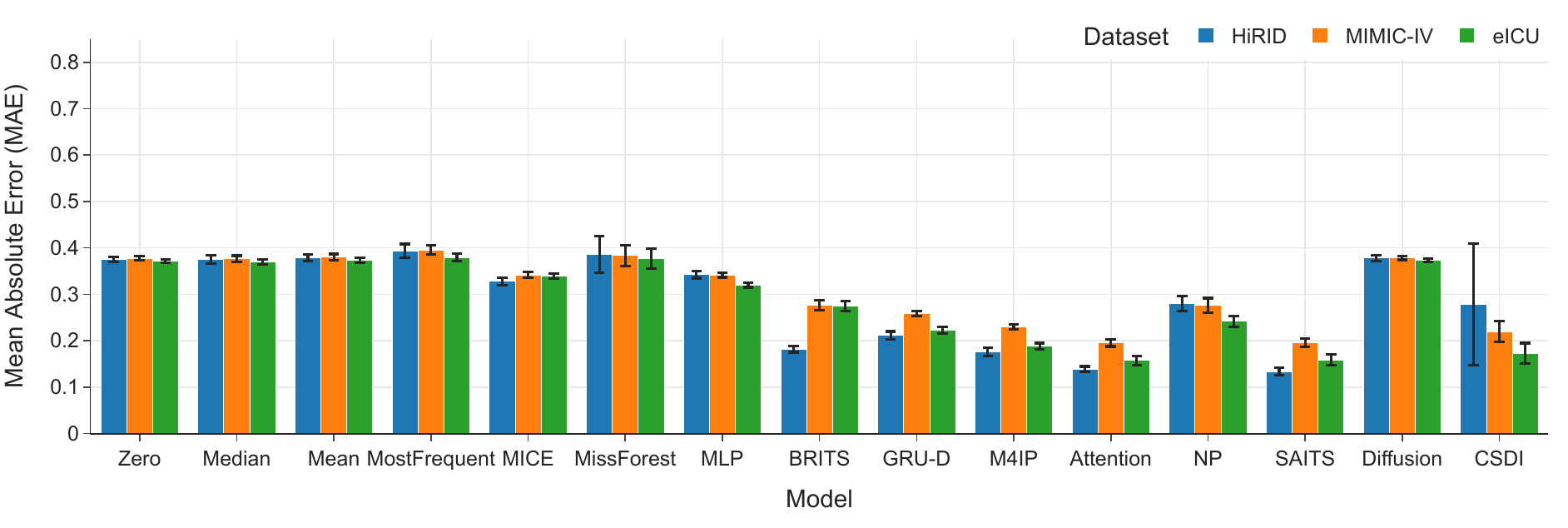}
    \caption{\textit{Performance in MAE across the selected imputation methods in three dimensions.} \textbf{Top:} imputation methods separated by missingness proportion for \acrshort{mnar}. \textbf{Middle:} aggregated performance per missingness type. \textbf{Bottom:} aggregated performance for each dataset.}
    \label{fig:results_perimputation}
    \vspace{-4mm}
\end{figure}
\begin{table}[]
\caption{\textit{The three best RNN, Attention, and Generative type imputation methods.} We \textbf{embolden} the best method per missingness type and metric including those within a standard deviation (±).}
\begin{tabularx}{\textwidth}{XX>{\centering\arraybackslash}X>{\centering\arraybackslash}X>{\centering\arraybackslash}X}
\toprule
                                          \textbf{Missingness}& \textbf{Metric} & \textbf{GRU-D}      & \textbf{Attention}  & \textbf{CSDI}      \\
                                          \hline
{\multirow{3}{*}{\textbf{MCAR}}} & RMSE   & 5.00±0.00      & 374.00±4.00 & \textbf{4.00±0.00}       \\
                      & MAE    & 0.18±0.00    & \textbf{0.14±0.00}     & 0.15±0.00    \\
                      & JSD    & \textbf{0.02±0.00}    & \textbf{0.02±0.00}    & \textbf{0.02±0.00}    \\
                      \hline
\multirow{3}{*}{\textbf{MAR}}                      & RMSE   & \textbf{5.00±0.00}       &372.00±32.00& 6.00±4.00       \\
                                          & MAE    & 0.19±0.02 & \textbf{0.13±0.01} & 0.18±0.08 \\ 
                                          & JSD    & 0.04±0.01 & \textbf{0.02±0.00}& 0.03±0.01 \\
                                          \hline
\multirow{3}{*}{\textbf{MNAR}}                     & RMSE   & \textbf{5.00±0.00}       & 419.00±27.00& \textbf{5.00±0.00}       \\
                                          & MAE    & 0.20±0.01  & \textbf{0.16±0.01}  & \textbf{0.17±0.01} \\
                                          & JSD    & 0.03±0.00    & \textbf{0.02±0.00}     & \textbf{0.02±0.00}    \\
                                          \hline
\multirow{3}{*}{\textbf{BO}}                       & RMSE   & \textbf{5.00±0.00}      & 485.00±30.00 & \textbf{5.00±1.00}       \\
                                          & MAE    & \textbf{0.18±0.00}    & 0.20±0.02   & 0.19±0.00    \\
                                          & JSD    & 0.03±0.00    & 0.03±0.01  & \textbf{0.02±0.00}   \\
\bottomrule
\end{tabularx}
\label{tab:results_per_category}
\end{table}
\subsection{Results}
We performed experiments with the introduced amputation methods, datasets, and imputation methods (\autoref{tab:imputation_methods}). 
\autoref{fig:results_perimputation} presents the results per imputation mechanism and the amount of missingness for MAE. Results for RMSE (\autoref{fig:results_perimputation_rmse}) and JSD (\autoref{fig:results_perimputation_jsd}) can be found in the appendix.

The top graph compares imputation methods for \acrlong{mnar}, a realistic missingness type for ICU data. This plot shows that, as expected, higher missingness is harder to impute and, additionally, results in higher standard deviations. Moreover, 70\% missingness doubles the MAE of 30\% . In the middle graph, we can conclude that \acrshort{bo} and \acrshort{mnar} are generally the hardest to impute. Moreover, CSDI has a high variance, which indicates mixed performance. The final plot shows HiRID is generally the easiest to impute, followed by eICU. In all these plots we see that Attention imputation slightly bests the other methods for MAE in each of the dimensions (missingness proportion, missingness type, and dataset).

We do not observe a decisive trend indicating that more recent models perform better; a relatively older model, GRU-D, seems to compete with more sophisticated models. 
\autoref{tab:results_per_category} further describes the performance of the three best imputation techniques per category for each type of missingness and performance metric. The results are similar for MAE and JSD, where the three methods are comparable. The Attention-based method slightly outperforms the other approaches. When it comes to RMSE, however, GRU-D and CSDI demonstrate significantly better performance; this could indicate that Attention imputation, along with several other methods (\autoref{fig:results_perimputation_rmse}), has a large error for the value of individual predictions. An algorithmic method is required - for example, we have no GPUs or explainability is required - MICE seems to be the best choice. Lastly, naive methods have comparable performance although the median imputation results in the best performance across all metrics.

\section{Discussion} 
Whereas newer methods often promise SOTA performance, the results depend on the type of task, and benchmarking may involve cherry-picking. RNN-type, attention, and generative models show promise for imputing time-series vital signs. The best technique depends on the type of missingness, the percentage of missingness, and the desired metric to minimize. We provide an openly accessible, extensible testbed to compare current and future imputation techniques on medical datasets of choice using YAIB. 

To make our comparison more robust, future work aims to include more diverse features, datasets, and types of imputation methods. Additionally, we recognize the importance of including downstream task performance as well as a more thorough discussion on the clinical applicability of imputation methods.
Our aim is to work towards a decision guide for \acrshort{ml} in health that can be used by clinicians and \acrshort{ml} researchers and increase common understanding.
\printbibliography
\section*{Author Contribution Statement}
Robin van de Water conceived the presented idea and supervised the project. Alisher Turubayev, Anna Shopova, Fabian Lange, Mahmut Kamalak, Paul Mattes and Victoria Ayvasky planned, executed and reported on experiments. Robin van de Water carried out the re-run of the experiments for this paper and prepared the original draft; Alisher Turubayev, Anna Shopova, Fabian Lange, Mahmut Kamalak, Paul Mattes and Victoria Ayvasky reviewed and revised the draft. 
\clearpage
\appendix
\section{Background and related work}

\begin{table}[h]
\small
\caption{\textit{Supplemental details of the considered ICU datasets.} Note that accessing each dataset requires completing a credentialing procedure.}
\label{tab:dataset_comparison_extended}
\centering
\begin{tabularx}{\textwidth}{l>{\centering\arraybackslash}m{0.30\textwidth}>{\centering\arraybackslash}m{0.20\textwidth}>{\centering\arraybackslash}m{0.20\textwidth}}
\toprule
\textbf{Dataset}                 & \textbf{\acrshort{miiii} / IV} & \textbf{\acrshort{eicu}}  & \textbf{\acrshort{hirid}}\\ 
\hline
\textbf{Stays*}              & 40k / 73k       & 201k     & 34k       \\   
\hline
\textbf{Version} & v1.4 / v2.2       & v2.0     & v1.1.1    \\       
\hline
\textbf{Frequency} & 1 hour    & 5 minutes & 2 / 5 minutes \\ 
\hline
\textbf{Origin}                 & USA       & USA       & Switzerland   \\ 
\hline
\textbf{Published} &
2015~\cite{johnsonMIMICIIIFreelyAccessible2016} / 2020~\cite{johnsonMIMICIVFreelyAccessible2023}& 2017~\cite{pollardEICUCollaborativeResearch2018}& 2020~\cite{hylandEarlyPredictionCirculatory2020} \\ 
\hline
\textbf{Benchmark}              &\cite{johnsonReproducibilityCriticalCare2017, purushothamBenchmarkingDeepLearning2018, harutyunyanMultitaskLearningBenchmarking2019, barbieriBenchmarkingDeepLearning2020, wangMIMICExtractDataExtraction2020,jarrettCLAIRVOYANCEPIPELINETOOLKIT2021,tangDemocratizingEHRAnalyses2020}/ \cite{xieBenchmarkingEmergencyDepartment2022}&~\cite{sheikhalishahiBenchmarkingMachineLearning2020, tangDemocratizingEHRAnalyses2020}&~\cite{yecheHiRIDICUBenchmarkComprehensiveMachine2022} \\ 
\hline
\textbf{Repository link} &\href{https://physionet.org/content/mimiciii/}{Physionet}/ \href{https://physionet.org/content/mimiciv/}{Physionet} &\href{https://physionet.org/content/eicu-crd/}{Physionet} & \href{https://physionet.org/content/hirid/}{Physionet} \\ 
\bottomrule
\end{tabularx}
\end{table}

\begin{table*}[h]
\begin{threeparttable}
    \centering
    \caption{\textit{Comparison table for the imputation benchmarks for medical time series.}}
    \small
    \begin{tabularx}{\textwidth}{llcccccccc}
        ~ & ~ &\rotatebox[origin=l]{90}{\textcite{jagerBenchmarkDataImputation2021}} & \rotatebox[origin=l]{90}{\textcite{sunReviewDeepLearning2020}} & \rotatebox[origin=l]{90}{\textcite{perez-lebelBenchmarkingMissingvaluesApproaches2022}} & \rotatebox[origin=l]{90}{\textcite{shadbahrClassificationDatasetsImputed2022}} & \rotatebox[origin=l]{90}{\textcite{psychogyiosMissingValueImputation2023}} & \rotatebox[origin=l]{90}{\textcite{luoEvaluatingStateArt2022}} & \rotatebox[origin=l]{90}{\textcite{jarrettHyperImputeGeneralizedIterative2022}} & \rotatebox[origin=l]{90}{\textbf{Our Work}} \\ \hline
        \textbf{Task} & Imputation & \cmark & \cmark &  \xmark & \cmark & \cmark & \cmark & \cmark & \textbf{\cmark} \\ 
         ~ & Downstream Task & \cmark & \cmark & \cmark & \cmark & \cmark &  \xmark &  \xmark & \textbf{\cmark} \\ \hline
        \textbf{Methods} & Naive & \cmark &  \xmark & \cmark & \cmark & \cmark & \cmark & \cmark & \textbf{\cmark} \\ 
         ~ & Algorithmic &  \xmark &  \xmark & \cmark & \cmark &  \xmark & \cmark & \cmark & \textbf{\cmark} \\ 
         ~ & Machine Learning & \cmark &  \xmark & \cmark & \cmark & \cmark & \cmark & \cmark & \textbf{\cmark} \\ 
         ~ & GAN-based & \cmark & \cmark &  \xmark & \cmark & \cmark &  \xmark & \cmark & \textbf{\cmark} \\ 
         ~ & RNN-based &  \xmark & \cmark &  \xmark &  \xmark &  \xmark & \cmark &  \xmark & \textbf{\cmark} \\ 
         ~ & AE-based & \cmark &  \xmark &  \xmark & \cmark & \cmark &  \xmark & \cmark &  \textbf{\cmark} \\ 
         ~ & Attention-based &  \xmark &  \xmark &  \xmark &  \xmark &  \xmark &  \xmark &  \xmark & \textbf{\cmark} \\ 
         ~ & Diffusion Models &  \xmark &  \xmark &  \xmark &  \xmark &  \xmark &  \xmark &  \xmark & \textbf{\cmark} \\ 
         ~ & Neural Processes &  \xmark &  \xmark &  \xmark &  \xmark &  \xmark &  \xmark &  \xmark & \textbf{\cmark} \\ 
         ~ & Available methods $^\P$ & -& - & 5 & - & - & 1 & 13 & \textbf{25} \\
         ~ & Benchmarked methods & 6 & 9 & 5 & 5 & 8 & 12 & 13 & \textbf{15} \\ \hline
        \textbf{Missingness} & MCAR & \cmark &  \xmark &  \xmark & \cmark & \cmark & \cmark & \cmark & \textbf{\cmark} \\ 
         ~ & MAR & \cmark &  \xmark &  \xmark &  \xmark &  \xmark &  \xmark & \cmark & \textbf{\cmark} \\ 
         ~ & MNAR & \cmark &  \xmark &  \xmark &  \xmark &  \xmark &  \xmark & \cmark & \textbf{\cmark} \\ 
         ~ & BO &  \xmark &  \xmark & \xmark & \xmark &  \xmark & \xmark &  \xmark & \textbf{\cmark} \\ 
        ~ & Native &  \xmark &  \xmark & \cmark & \xmark &  \xmark & \cmark &  \xmark & \textbf{\cmark} \\
         \hline
        \textbf{Datasets} & MIMIC$^\dag$ &  \xmark & III & III & III &  \xmark & III &  \xmark & \textbf{III/IV} \\ 
         ~ & eICU &  \xmark &  \xmark &  \xmark &  \xmark &  \xmark &  \xmark &  \xmark & \textbf{\cmark} \\ 
         ~ & HIRID &  \xmark &  \xmark &  \xmark &  \xmark &  \xmark &  \xmark &  \xmark & \textbf{\cmark} \\ 
         ~ & AUMCdb &  \xmark &  \xmark &  \xmark &  \xmark &  \xmark &  \xmark &  \xmark & \textbf{\cmark} \\ 
         ~ & Other medical & \cmark & \cmark & \cmark & \cmark & \cmark &  \xmark & \cmark &  \textbf{\cmark} \\ 
         \hline
        \textbf{Functionality} & Hyperparameter Tuning & \cmark &  \xmark &  \xmark & \cmark &  \xmark &  \xmark &  \xmark & \textbf{\cmark} \\ 
         ~ & Code Availability &  \xmark &  \xmark & \cmark &  \xmark &  \xmark & \cmark & \cmark & \textbf{\cmark} \\ 
         ~ & Time-Series Data &  \xmark & \cmark &  \xmark &  \xmark &  \xmark &  \xmark & \cmark & \textbf{\cmark} \\ 
         ~ & Data Amputation &  \xmark &  \xmark & \xmark &  \xmark &  \xmark & \xmark & \xmark & \textbf{\cmark} \\ 
         ~ & Extensibility &  \xmark &  \xmark & \xmark &  \xmark &  \xmark & \xmark & \cmark & \textbf{\cmark} \\ 
         \hline
    \end{tabularx}
    \label{tab:related-works-overview}
\end{threeparttable}
\end{table*}

\clearpage
\section{Extended Results}
Note that we aggregate the means and standard deviations over several different runs. We aim to get a comprehensive result summarization with this method.

\begin{figure}[h]
    \centering
    \centering
    \includegraphics[width=\textwidth]{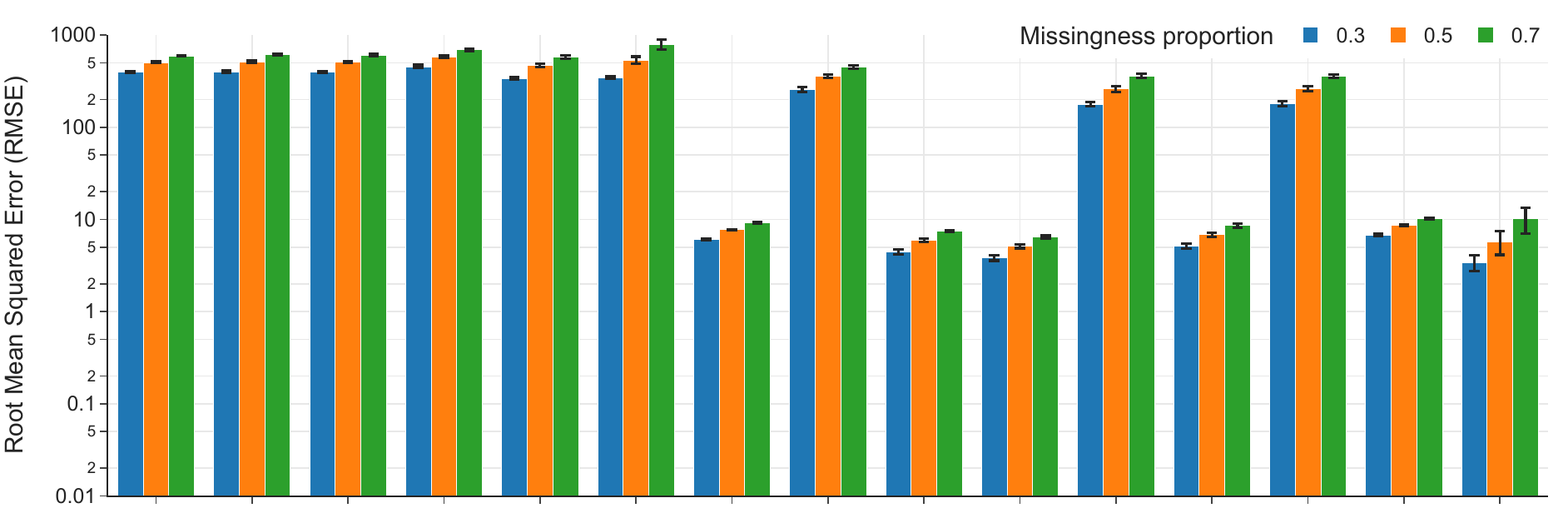}
    \includegraphics[width=\textwidth]{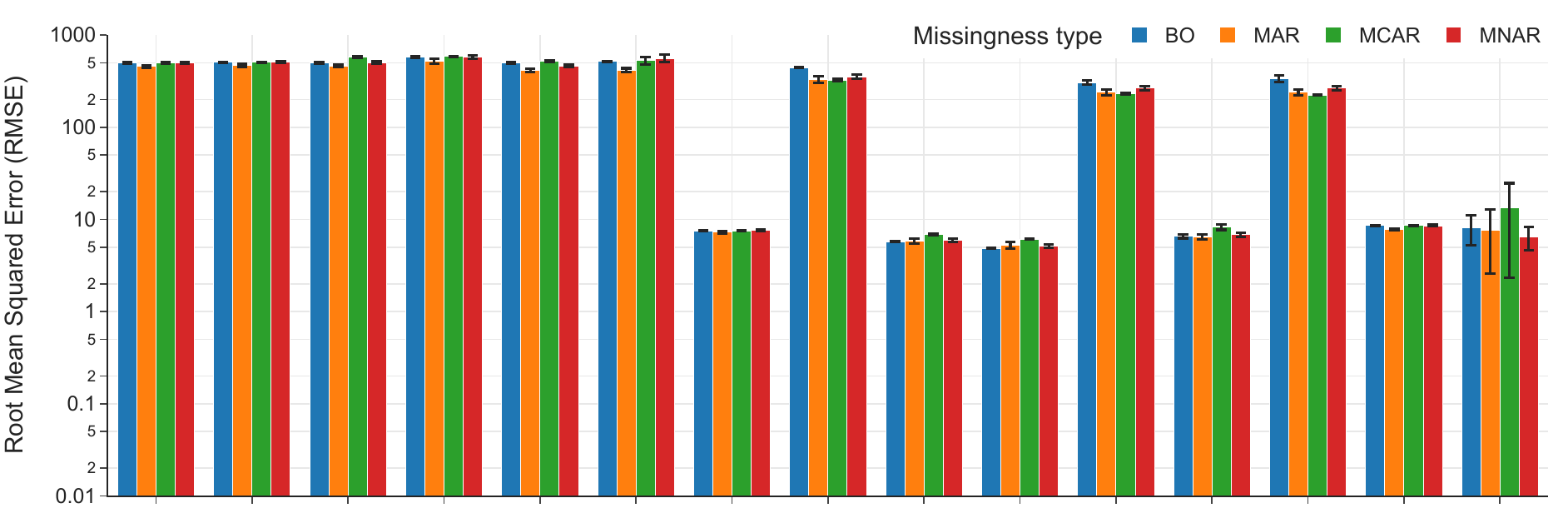}
    \includegraphics[width=\textwidth]{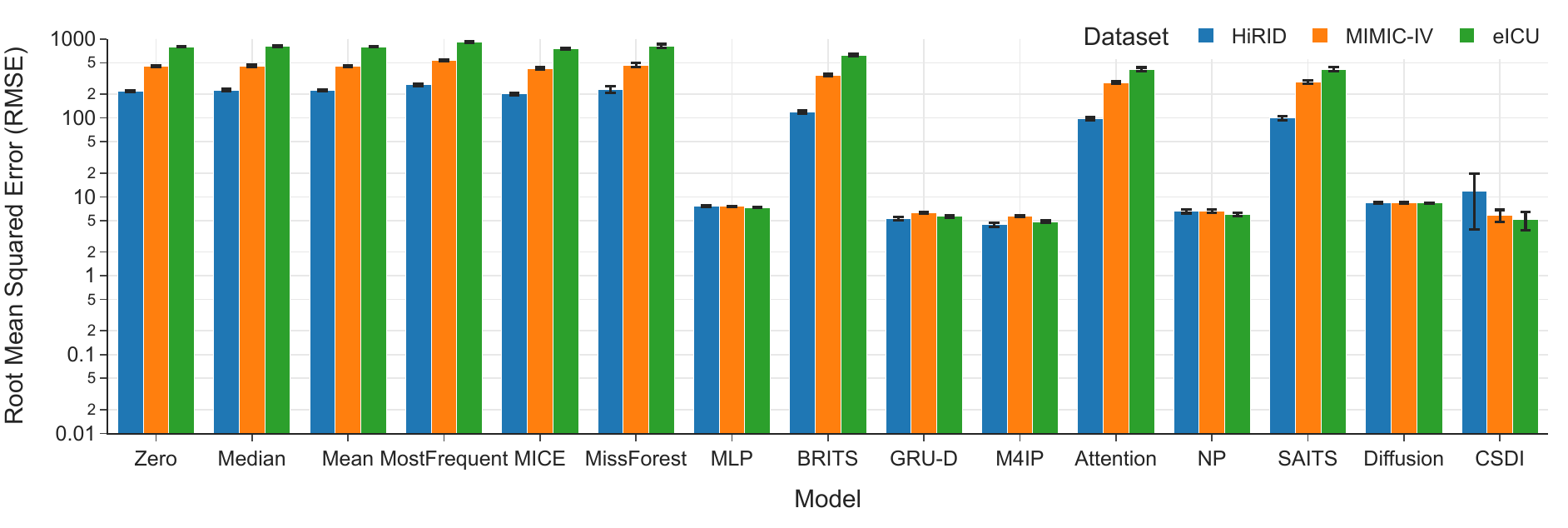}
    \caption{\textit{Performance in RMSE across the selected imputation methods in three dimensions.} Note that we use a log scale for readability. \textbf{Top:} imputation methods separated by missingness proportion for \acrshort{mnar}. \textbf{Middle:} aggregated performance per missingness type. \textbf{Bottom:} aggregated performance for each dataset.}
    \label{fig:results_perimputation_rmse}
\end{figure}

\begin{figure}[p]
    \centering
    \centering
    \includegraphics[width=\textwidth]{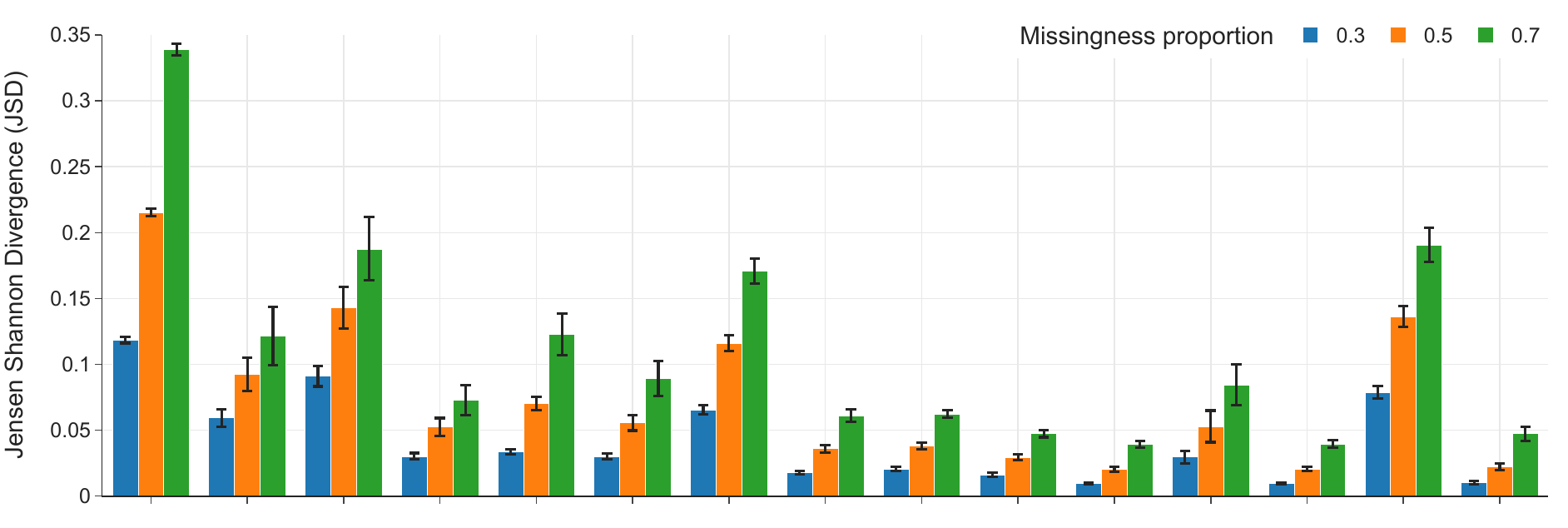}
    \includegraphics[width=\textwidth]{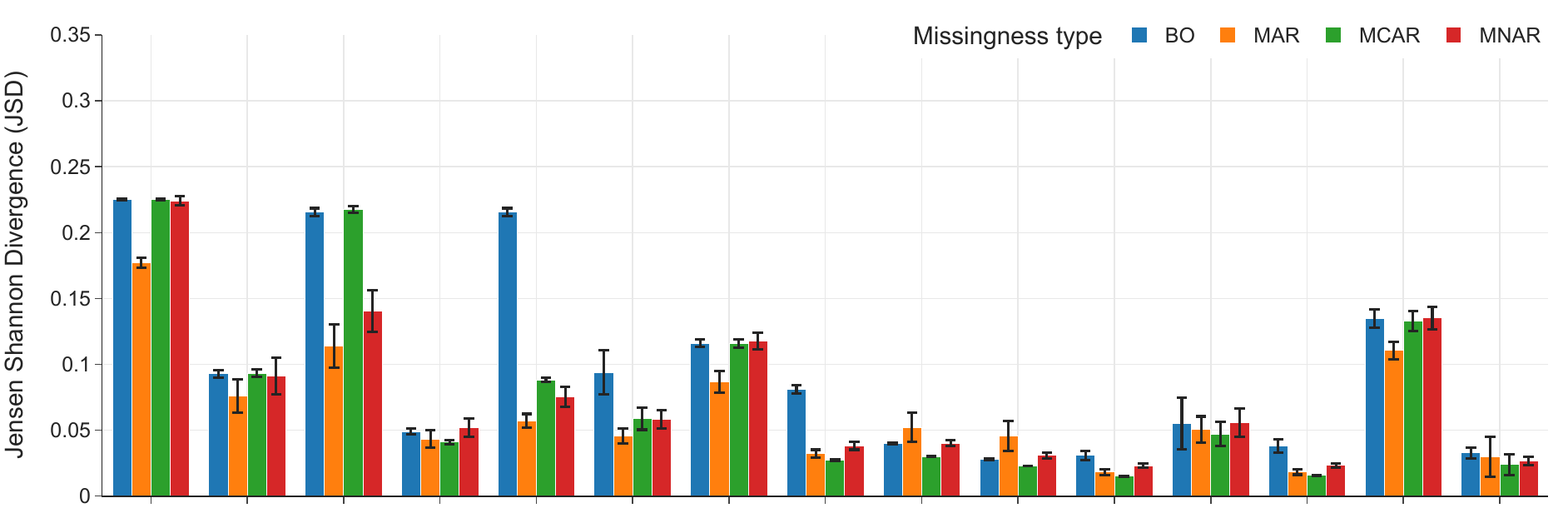}
    \includegraphics[width=\textwidth]{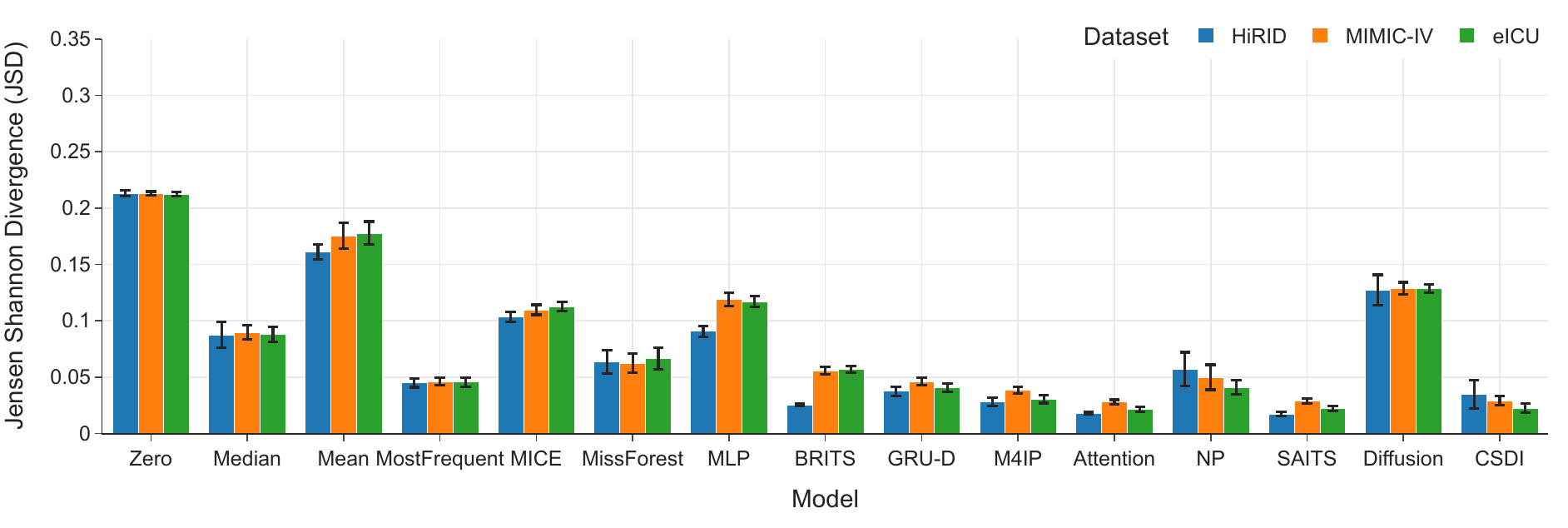}
    \caption{\textit{Performance in JSD across the selected imputation methods in three dimensions.} \textbf{Top:} imputation methods separated by missingness proportion for \acrshort{mnar}. \textbf{Middle:} aggregated performance per missingness type. \textbf{Bottom:} aggregated performance for each dataset.}
    \label{fig:results_perimputation_jsd}
\end{figure}

\autoref{fig:results_perimputation_rmse} and \autoref{fig:results_perimputation_jsd} display the results in the same manner as \autoref{fig:results_perimputation} in the main text, but for RMSE and JSD.

\autoref{tab:missingness_type_dataset} shows the missingness results averaged over each dataset (row) and missingness mechanism (grouped columns) for every metric (individual columns). \autoref{tab:results_perimputation} shows the RMSE and MAE for each imputation metric over missingness quantities.

\begin{table}[p]
\setlength{\tabcolsep}{3pt}
\small
\caption{\textit{Results of all benchmarked imputation methods aggregated by missingness type averaged by dataset.} }
\begin{tabularx}{\textwidth}{l>{\centering\arraybackslash}X>{\centering\arraybackslash}X>{\centering\arraybackslash}X>{\hsize=.05\hsize}X>{\centering\arraybackslash}X>{\centering\arraybackslash}X>{\centering\arraybackslash}X>{\hsize=.05\hsize}X>{\centering\arraybackslash}X>{\centering\arraybackslash}X>{\centering\arraybackslash}X>{\hsize=.05\hsize}X>{\centering\arraybackslash}X>{\centering\arraybackslash}X>{\centering\arraybackslash}X}
\toprule
\textbf{Type} & \multicolumn{3}{c}{\textbf{MCAR}}& &\multicolumn{3}{c}{\textbf{MAR}}&  & \multicolumn{3}{c}{\textbf{MNAR}}         &  & \multicolumn{3}{c}{\textbf{BO}}             \\
\cline{2-4}
\cline{6-8}
\cline{10-12}
\cline{14-16}

\addlinespace[0.2em]
    & \textbf{RMSE} & \textbf{MAE} & \textbf{JSD} & & \textbf{RMSE} & \textbf{MAE} & \textbf{JSD} & & \textbf{RMSE} & \textbf{MAE} & \textbf{JSD} & &\textbf{RMSE} & \textbf{MAE} & \textbf{JSD} \\
\hline
\underline{\textbf{Dataset}}\\
\textbf{MIIV}       & 325           & 0.32         & \textbf{0.07}    & & 305           & 0.30         & \textbf{0.06}        & & 362          & 0.38         & 0.08       &  & 505           & 0.52         & \textbf{0.10}      \\
\textbf{eICU}       & 604           & 0.33         & \textbf{0.07}    & & 527           & \textbf{0.28}         & \textbf{0.06}        & & 634          & \textbf{0.36}         & \textbf{0.07}       &  & 978           & 0.53         & \textbf{0.10}         \\
\textbf{HiRID}      & \textbf{159}           & \textbf{0.32}         & \textbf{0.07 }   & & \textbf{142}           & \textbf{0.28}         & \textbf{0.06}        & & \textbf{171}          & \textbf{0.36}         & \textbf{0.07}        & & \textbf{244}           & \textbf{0.50}         & \textbf{0.10}         \\

\bottomrule
\end{tabularx}
\label{tab:missingness_type_dataset}
\end{table}
\begin{table}[h]
\caption{\textit{Base results averaged over four amputation mechanisms and three datasets, grouped by missingness level (30\%, 50\%, 70\%)}. We \textbf{embolden} the best model per column and those within a standard deviation (±). RMSE: Root Mean Squared Error ($\downarrow$,  i.e., lower is better), MAE: Mean Absolute Error ($\downarrow$)}
\begin{tabularx}{\textwidth}{l>{\centering\arraybackslash}X>{\centering\arraybackslash\leavevmode\color{darkgray!60}}X>{\centering\arraybackslash}X>{\centering\arraybackslash\leavevmode\color{darkgray!60}}X>{\centering\arraybackslash}X>{\centering\arraybackslash\leavevmode\color{darkgray!60}}X}
\toprule
\multicolumn{1}{l|}{}  & \multicolumn{6}{c|}{\textbf{Missingness Quantity}}                                               \\ \cline{2-7}
\multicolumn{1}{|c|}{} & \multicolumn{2}{c|}{\textbf{30\%}} & \multicolumn{2}{c|}{\textbf{50\%}} & \multicolumn{2}{c|}{\textbf{70\%}} \\ \hline
\addlinespace[0.2em]
\textbf{Model}                  & \textbf{RMSE}        & \textbf{MAE}        & \textbf{RMSE}         & \textbf{MAE}       & \textbf{RMSE}        & \textbf{MAE}        \\
\hline
Zero                   & 386.6±8     & 0.23±0.00     & 496.4±8.3    & 0.37±0.00    & 586.1±8.7   & 0.52±0.01 \\
Median                 & 391.6±10.2  & 0.22±0.01  & 503.9±11     & 0.37±0.01 & 597.6±12.2  & 0.52±0.01  \\
Mean                   & 387.4±8.6   & 0.23±0.00     & 498.8±9.6    & 0.38±0.01 & 635.7±12.4  & 0.53±0.01  \\
MostFrequent           & 445.7±15.2  & 0.25±0.01  & 573±16.8     & 0.41±0.01 & 678.6±18.8  & 0.58±0.02  \\
\hline
MICE                   & 337.5±8.5   & 0.18±0.00     & 467.7±12.5   & 0.33±0.01 & 611.4±15.8  & 0.49±0.01  \\
MissForest             & 349.1±5.8   & 0.19±0.00     & 497.3±23.1   & 0.36±0.02 & 672.7±62.6  & 0.59±0.06  \\
\hline
MLP                    & 5.9±0.1     & 0.20±0.00      & 7.6±0.1      & 0.33±0.01 & 9.0±0.1       & 0.47±0.01  \\
BRITS                  & 261.6±13.2  & 0.13±0.01  & 368.2±14.2   & 0.24±0.01 & 458.3±15.9  & 0.34±0.01  \\
GRU-D                   & \textbf{3.7±0.2}     & 0.11±0.00     & \textbf{5.0±0.2}        & 0.19±0.01 & \textbf{7.2±0.2}     & 0.28±0.01  \\
M4IP                    & 4.3±0.2     & 0.13±0.00     & 5.8±0.2      & 0.23±0.01 & 8.1±0.2     & 0.33±0.01  \\
\hline
Attention              & 179.4±7.1   & \textbf{0.08±0.00}     & 258.1±11.4   & \textbf{0.15±0.01} & 343±18.6    & \textbf{0.24±0.01}  \\
NP                     & 4.9±0.2     & 0.16±0.01  & 6.6±0.3      & 0.28±0.02 & 10.1±0.6    & 0.41±0.02  \\
SAITS                  & 180.8±8.8   & \textbf{0.08±0.00}     & 267.7±17.5   & \textbf{0.16±0.01} & 369.1±22    & 0.28±0.02  \\
\hline
Diffusion              & 6.6±0.1     & 0.23±0.00     & 8.5±0.1      & 0.38±0.00    & 10.0±0.1      & 0.53±0.01 \\
CSDI                   & 4.0±1.6     & 0.10±0.02   & 6.3±2.5      & 0.19±0.03 & 18.0±13.2     & 0.37±0.14  \\
\bottomrule
\end{tabularx}
\label{tab:results_perimputation}
\end{table}

\clearpage
\section{Data characteristics}
We have analyzed the datasets we used in this work to discover patterns that might be interesting for imputation performance. 

\begin{figure}[ht!]
\begin{subfigure}{.48\linewidth}
  \includegraphics[width=\linewidth]{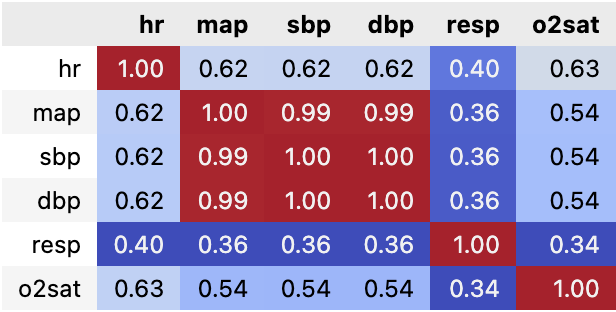}
  \caption{eICU}
  \label{eicu-corr}
\end{subfigure}\hfill 
\begin{subfigure}{.48\linewidth}
  \includegraphics[width=\linewidth]{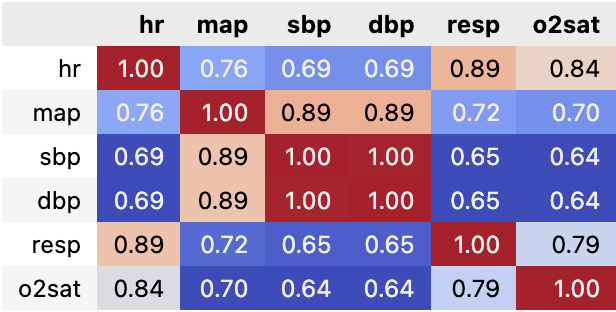}
  \caption{MIMIC-IV}
  \label{mimic-corr}
\end{subfigure}
\medskip 
\begin{center}    
\begin{subfigure}{.48\linewidth}
  \includegraphics[width=\linewidth]{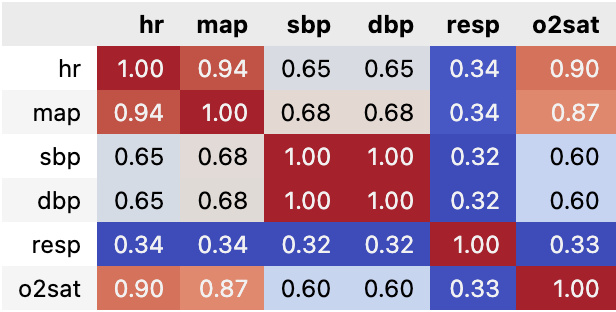}
  \caption{HiRiD}
  \label{hirid-corr}
\end{subfigure}
\end{center}
\caption{\textit{Missingness correlation of the selected features for each dataset.}}
\label{fig:correlation}
\end{figure}

\autoref{fig:correlation} shows the missingness correlation of the features of each dataset. We can see that some features are heavily missing at the same timestep, depending on the dataset. This can have impact on the result of multiple imputation methods. 

\begin{figure}[ht!]
\begin{subfigure}{.33\linewidth}
  \includegraphics[width=\linewidth]{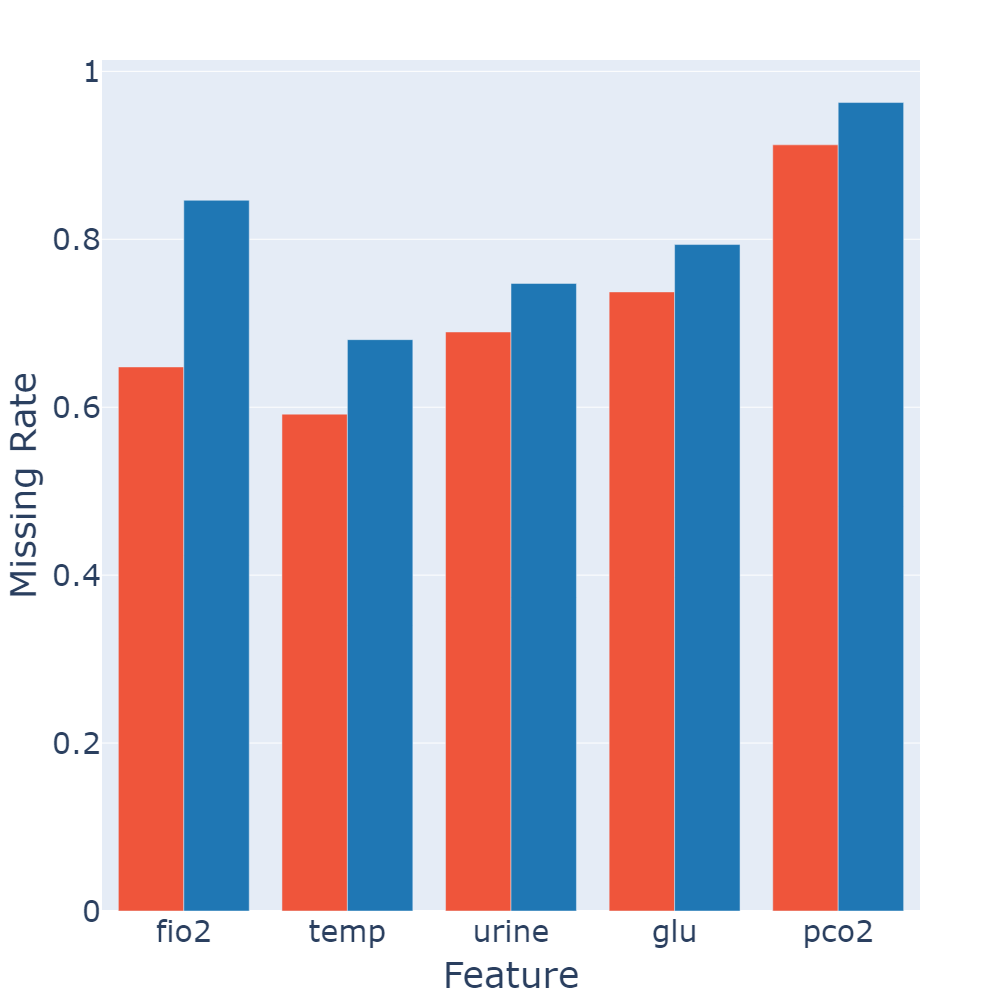}
  \caption{eICU}
  \label{eicu-inf-missingness}
\end{subfigure}\hfill 
\begin{subfigure}{.33\linewidth}
  \includegraphics[width=\linewidth]{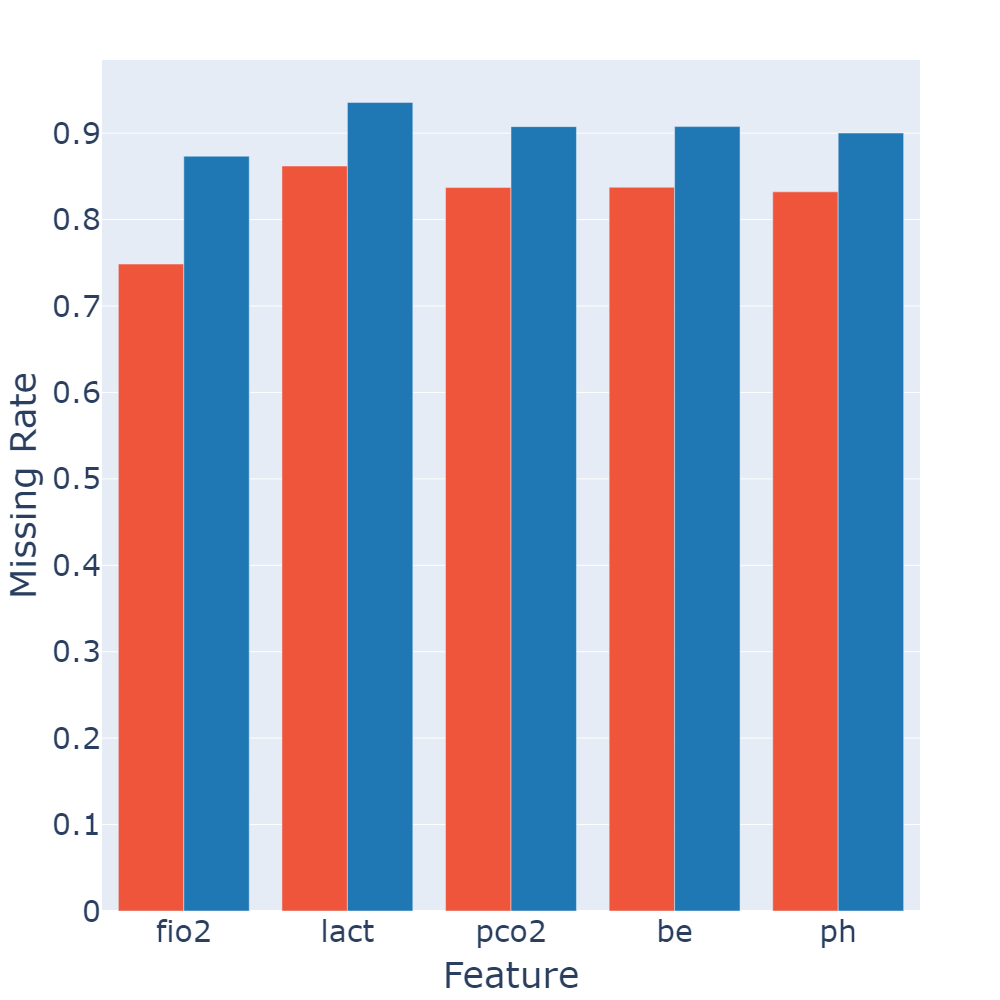}
  \caption{MIMIC-IV}
  \label{mimic-inf-missingness}
\end{subfigure}
\medskip
\centering
\begin{subfigure}{.33\linewidth}
  \includegraphics[width=\linewidth]{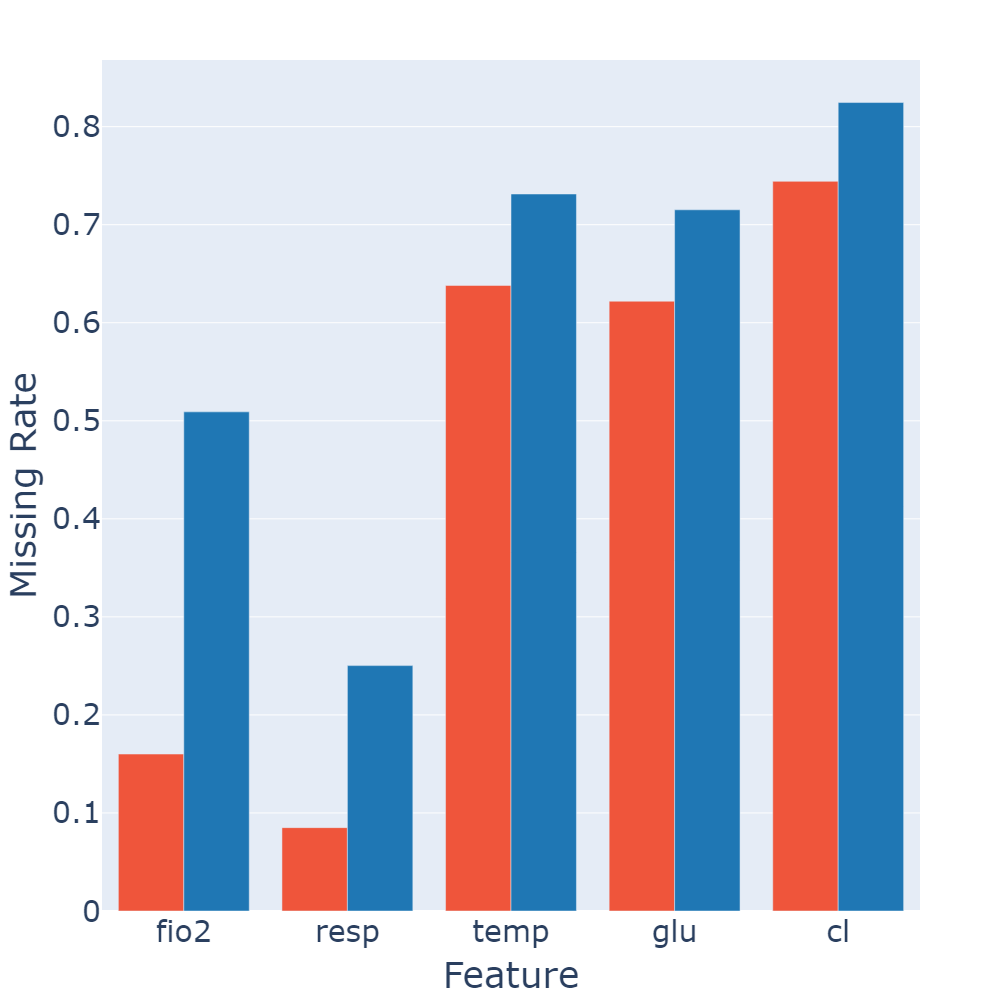}
  \caption{HiRiD}
  \label{hirid-inf-missingness}
\end{subfigure}\hfill 
\caption{\textit{Missingness rate for five features with the highest difference in missing rates between the classes survivor (blue) and non-survivor (red) in each dataset.}}
\label{fig:informative-missingness}
\end{figure}

\autoref{fig:informative-missingness} explores the concept of informative missingness: we observe a generally higher missingness for survivors than for non-survivors, although this difference is not dramatic. We might ascribe this to a clinical decision to monitor patients in a worse state more frequently.


\clearpage
\section{Extensibility}
\label{app:extensibility}
We have added a range of imputation methods to \acrshort{yaib}~\cite{vandewaterAnotherICUBenchmark2023}, including interfaces to existing imputation libraries~\cite{duPyPOTSPythonToolbox2023, jarrettHyperImputeGeneralizedIterative2022}. Here, we describe the addition of a recently introduced method that uses conditional score-based diffusion models conditioned on observed data: the \ac{csdi}\cite{tashiroCSDIConditionalScorebased2021}. To make the process of implementing these models easier, we have created the \texttt{ImputationWrapper} class that extends the pre-existing \texttt{DLWrapper} (itself a subclass of the \texttt{LightningModule} of Pytorch-lightning) with extra functionality. 

The \acrshort{csdi} model is a diffusion model that follows the general architecture of conditional diffusion models~\cite{hoDenoisingDiffusionProbabilistic2020}; It introduces noise into a subset of time series data used as conditional observations to later denoise the data and predict accurate values for the imputation targets. \ac{csdi} is based on a U-Net architecture\cite{ronnebergerUNetConvolutionalNetworks2015} including residual connections.

\begin{figure}[h]
\centering
\includegraphics[width=0.8\textwidth]{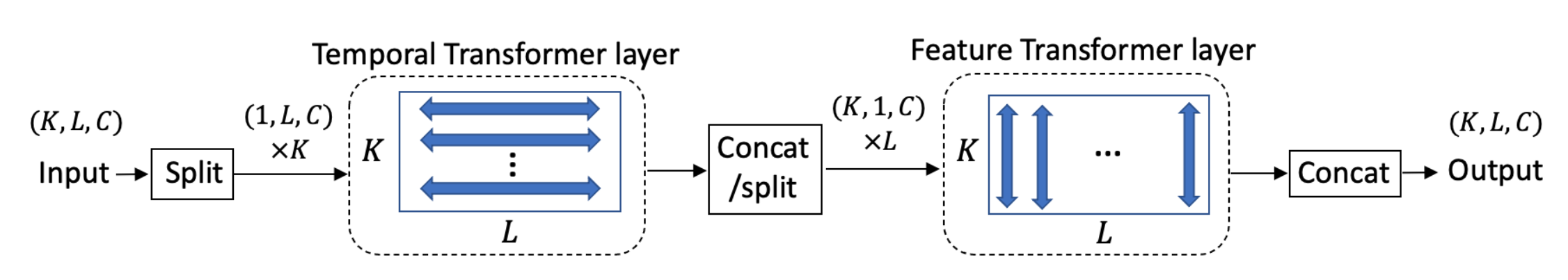}
\caption{The attention mechanism of \ac{csdi} adapted from \cite{tashiroCSDIConditionalScorebased2021}.}
\label{fig:csdi_attention}
\end{figure}

\cite{tashiroCSDIConditionalScorebased2021} included two additional features into their model, which are inspired by DiffWave~\cite{kongDiffWaveVersatileDiffusion2021}: an attention mechanism and the ability to input side information. The attention mechanism uses transformer layers, as shown in \autoref{fig:csdi_attention}. An input with K features, L length, and C channels is reshaped first to apply temporal attention and later reshaped again to apply feature attention. The second additional feature allows side information to be used as input to the model by a categorical feature embedding \cite{tashiroCSDIConditionalScorebased2021}.

See \autoref{code:csdi-init} for the most important implementation code: the model initialization. We note that of this code, very little has been adapted from the original code repository\footnote{\url{https://github.com/ermongroup/CSDI/tree/main}} included in the original publication~\cite{tashiroCSDIConditionalScorebased2021}.

\begin{lstlisting}[frame=single, float=h, style=pycharm, language=Python, caption={\textit{Implementing the CSDI architecture in YAIB}. Note that %\href{https://github.com/rvandewater/YAIB/blob/development/icu_benchmarks/imputation/csdi.py}{our implementation}
our implementation is very similar to the \href{https://github.com/ermongroup/CSDI/tree/main}{original github repository}, which demonstrates the flexibility of implementing new models in YAIB.}, columns=fullflexible, label=code:csdi-init, basicstyle=\ttfamily\tiny]
{
    def __init__(
        self, input_size, time_step_embedding_size, feature_embedding_size, unconditional, target_strategy, num_diffusion_steps, diffusion_step_embedding_dim, n_attention_heads, num_residual_layers, noise_schedule, beta_start, beta_end, n_samples, conv_channels, *args, **kwargs,
    ):
        super().__init__(...) 
        self.target_dim = input_size[2]
        self.n_samples = n_samples

        self.emb_time_dim = time_step_embedding_size
        self.emb_feature_dim = feature_embedding_size
        self.is_unconditional = unconditional
        self.target_strategy = target_strategy

        self.emb_total_dim = self.emb_time_dim + self.emb_feature_dim
        if not self.is_unconditional:
            self.emb_total_dim += 1  # for conditional mask
        self.embed_layer = nn.Embedding(num_embeddings=self.target_dim, embedding_dim=self.emb_feature_dim)

        input_dim = 1 if self.is_unconditional else 2
        self.diffmodel = diff_CSDI(
            conv_channels,
            num_diffusion_steps,
            diffusion_step_embedding_dim,
            self.emb_total_dim,
            n_attention_heads,
            num_residual_layers,
            input_dim,
        )

        # parameters for diffusion models
        self.num_steps = num_diffusion_steps
        if noise_schedule == "quad":
            self.beta = np.linspace(beta_start**0.5, beta_end**0.5, self.num_steps) ** 2
        elif noise_schedule == "linear":
            self.beta = np.linspace(beta_start, beta_end, self.num_steps)

        self.alpha_hat = 1 - self.beta
        self.alpha = np.cumprod(self.alpha_hat)
        self.alpha_torch = torch.tensor(self.alpha).float().unsqueeze(1).unsqueeze(1)

}
\end{lstlisting}

\clearpage
\section{Hyperparameters}
\label{sec:hyperparameter_optimization_tables}
In the following section, we list the hyperparameters we investigated and the values that were determined best and were used in the final evaluation.

\begin{table}[h]
\centering
\begin{tabular}{|l|c|c|}
\hline
Optimized Parameter       & Possible Values  & Chosen Value \\
\hline
Adam Learning Rate        & 0.001, 0.01, 0.1 & 0.1          \\
Learning Rate Scheduler   & Cosine, None     & Cosine       \\
Number of Layers          & 2, 4, 6          & 4            \\
$d_{model}$              & 64, 128, 256     & 256          \\
$d_{inner}$             & 64, 128, 256     & 128          \\
Number of Attention Heads & 4, 8, 16         & 4            \\
$d_k$                      & 32, 64, 128      & 32           \\
$d_v$                      & 32, 64, 128      & 32           \\
Dropout Percentage        & 0.0, 0.1, 0.3    & 0.0         \\
\hline
\end{tabular}
\label{tab:attention_hyperparameters}
\caption{Optimized Hyperparameters for the Attention Imputation Model}
\end{table}

\begin{table}[h]
\centering
\begin{tabular}{|l|c|c|}
\hline
Optimized Parameter            & Possible Values       & Chosen Value \\
\hline
Adam Learning Rate             & 0.001, 0.01, 0.1      & 0.001          \\
Learning Rate Scheduler        & Cosine, None          & None       \\
Cell Type                      & GRU, LSTM             & LSTM         \\
Hidden Size                    & 32, 64, 128           & 64           \\
Initialization of Hidden State & Zeros, Gaussian Noise & Zero         \\
Dropout Percentage             & 0.0, 0.1, 0.3         & 0.0         \\
\hline
\end{tabular}
\label{tab:brnn_hyperparameters}
\caption{Optimized Hyperparameters for the 
D-GRU Imputation Model}
\end{table}

\begin{table}[h]
\centering
\begin{tabular}{|l|c|c|}
\hline
Optimized Parameter            & Possible Values       & Chosen Value \\
\hline
Adam learning rate                 & 0.001, 0.01, 0.1     & 0.001        \\
learning rate scheduler            & cosine, None         & cosine       \\
time step embedding size           & 64, 128, 256         & 64           \\
feature embedding size             & 8, 16, 32, 64        & 64           \\
target strategy                    & random, history, mix & history      \\
number of diffusion steps          & 50, 100, 200         & 50           \\
diffusion step embedding dimension & 64, 128, 256         & 128          \\
number of attention heads          & 6, 8, 10             & 8            \\
noise schedule                     & quadratic, linear    & quadratic    \\
number of diffusion samples        & 5, 10, 15            & 15           \\
number of residual layers          & 2, 4, 6, 8           & 8           \\
\hline
\end{tabular}
\label{tab:csdi_hyperparameters}
\caption{Optimized Hyperparameters for the CSDI Imputation Model}
\end{table}

\begin{table}[h!]
\centering
\begin{tabular}{|l|c|c|}
\hline
Optimized Parameter            & Possible Values       & Chosen Value \\
\hline
Adam learning rate       & 0.001, 0.01, 0.1 & 0.1          \\
learning rate scheduler  & cosine, None     & cosine       \\
encoder layers           & 3, 6, 12         & 6            \\
encoder hidden dimension & 24, 36, 72       & 72           \\
decoder layers           & 3, 6, 12         & 3            \\
decoder hidden dimension & 24, 36, 72       & 72           \\
r dimension              & 3, 6, 12         & 12           \\
z dimension              & 3, 6, 12         & 12           \\
\hline
\end{tabular}
\label{tab:np_hyperparameters}
\caption{Optimized Hyperparameters for the Neural Process Imputation Model}
\end{table}

\begin{table}[h]
\centering
\begin{tabular}{|l|c|c|}
\hline
Optimized Parameter            & Possible Values       & Chosen Value \\
\hline
Adam learning rate                 & 0.001, 0.01, 0.1     & 0.001        \\
learning rate scheduler            & cosine, None         & cosine       \\
number of layers          & 2, 4, 6          & 4            \\
$d_{model}$              & 64, 128, 256     & 128          \\
$d_{inner}$             & 64, 128, 256     & 128          \\
number of attention heads & 4, 8, 16         & 4            \\
$d_k$                      & 32, 64, 128      & 32           \\
$d_v$                      & 32, 64, 128      & 32           \\
dropout percentage        & 0.0, 0.1, 0.3    & 0.1         \\
\hline
\end{tabular}
\label{tab:saits_hyperparameters}
\caption{Optimized Hyperparameters for the SAITS Imputation Model}
\end{table}

\begin{table}[h]
\centering
\begin{tabular}{|l|c|c|}
\hline
Optimized Parameter            & Possible Values       & Chosen Value \\
\hline
Adam learning rate                 & 0.001, 0.01, 0.1     & 0.1        \\
learning rate scheduler            & cosine, None         & None       \\
residual channel size           & 64, 128, 256         & 64           \\
skip connection channel size             & 64, 128, 256        & 256           \\
number of residual layers          & 12, 24, 36         & 36           \\
diffusion step embedding dimension input & 64, 128, 256 & 256 \\ 
diffusion step embedding dimension middle & 64, 128, 256 & 128 \\ 
diffusion step embedding dimension output & 64, 128, 256 & 256 \\
S4 max length & 50, 100, 200 & 100 \\
S4 state dimension & 32, 64, 128 & 64 \\
S4 dropout & 0.0, 0.1, 0.3 & 0.3 \\
S4 bidirectional & true, false & false \\
S4 layer normalization & true, false & false \\
diffusion time steps & 500, 1000, 2000 & 2000 \\
\hline
\end{tabular}
\label{tab:sssds4_hyperparameters}
\caption{Optimized Hyperparameters for the SSSDS4 Imputation Model}
\end{table}

\end{document}